\documentclass{llncs}
\usepackage{llncsdoc}
\usepackage{graphicx}
\begin{document}
\markboth{\LaTeXe{} Class for Lecture Notes in Computer
Science}{\LaTeXe{} Class for Lecture Notes in Computer Science}
\thispagestyle{empty}

\newpage
\newcounter{save}\setcounter{save}{\value{section}}
{\def\addtocontents#1#2{}%
\def\addcontentsline#1#2#3{}%
\def\markboth#1#2{}%
\title{Zoom-in-Net: Deep Mining Lesions for Diabetic Retinopathy Detection}

\author{Zhe Wang\inst{1}, Yanxin Yin\inst{2}, Jianping Shi\inst{3}, Wei Fang\inst{4}, Hongsheng Li\inst{1} \and Xiaogang Wang\inst{1}  }

\institute{The Chinese University of Hong Kong, Hong Kong\and Tsing Hua University, Beijing, China \and SenseTime Group Limited, Beijing, China \and Sir Run Run Shaw Hospital, Hangzhou, China}

\maketitle
\begin{abstract}

We propose a convolution neural network based algorithm for simultaneously diagnosing diabetic retinopathy and highlighting suspicious regions. Our contributions are two folds: 1) a network termed Zoom-in-Net which mimics the zoom-in process of a clinician to examine the retinal images. Trained with only image-level supervisions, Zoom-in-Net can generate attention maps which highlight suspicious regions, and predicts the disease level accurately based on both the whole image and its high resolution suspicious patches. 2) Only four bounding boxes generated from the automatically learned attention maps are enough to cover 80\% of the lesions labeled by an experienced ophthalmologist, which shows good localization ability of the attention maps. By clustering features at high response locations on the attention maps, we discover meaningful clusters which contain potential lesions in diabetic retinopathy.  Experiments show that our algorithm outperform the state-of-the-art methods on two datasets, EyePACS and Messidor.
\end{abstract}
\section{Introduction}

Identifying suspicious region for medical images is of significant importance since it provides intuitive illustrations for physicians and patients of how the diagnosis is made.  
However, most previous works rely on strong supervisions which require lesion location information. This largely limits the size of the dataset as the annotations in medical imaging are expensive to acquire. Therefore, it is necessary to develop algorithms which can make use of large datasets with weak supervisions for simultaneous classification and localization. 

In this work, we propose a general weakly supervised  learning framework, Zoom-in-Net, based on convolution neural networks (CNN). The proposed method is accurate in classification and meanwhile, it can also automatically discover the lesions in the images at a high recall with only several bounding boxes. This framework can be easily extended to various classification problems and provides convenient visual inspections for the doctors. 

To verify the effectiveness of our method, we aim to solve the problem of Diabetic retinopathy (DR) detection as it is an important problem and a large scale dataset \cite{drdlb} with image-level labels are publicly available online.
DR is an eye disease caused by diabetes. 
Today, it is the leading cause of blindness in the working-age population of the developed world.  Treatments can be applied to slow down or stop further vision loss once this disease is diagnosed. However, DR has no early warning sign and the diagnosis is a time-consuming and manual process that requires an experienced clinician to examine the retinal image. It is often too late to provide effective treatments because of the delay.
In order to alleviate the workload of human interpretation, various image analysis algorithms have been proposed over the last few decades. 

Early approaches \cite{chaum2008automated,abramoff2010automated} use hand-crafted features to represent the images, of which the main bottlenecks are the limited expressive power of the features.
Recently, CNN based methods \cite{chandrakumar2016classifying,abramoff2016improved,gulshan2016development} have dramatically improved the performance of DR detection.  Most of them treat CNN as a black box, which lacks intuitive explanation. 
Few previous works localize the lesions with image-level supervisions, such as visualizing the evidence hotspots of the spinal MRIs classification \cite{jamaludin2016spinenet}.
But they did not use the hotspots to further improve performance. 

The proposed Zoom-in-Net has the attention mechanism, which generates attention maps using only image-level supervisions. The attention map is a heatmap that indicates which pixels play more important roles in making the image-level decision. In addition, our Zoom-in-Net mimics the clinicians' behavior which skim the DR images to identify suspicious regions and then zoom-in to verify. 
Zoom-in-Net is validated on EyePACS and Messidor datasets. 
It outperforms state-of-the-art methods as well as general physicians. Moreover, 
we also validated the attention localization accuracy on around 200 images labeled by an experienced ophthalmologist. Our attention localization reaches a recall of 0.82 which proves to be useful for doctors. The clustered regions at high response locations of the attention maps shows meaningful lesions in diabetic retinopathy.

\begin{figure}[t]
\centering
  \includegraphics[height=5.9cm]{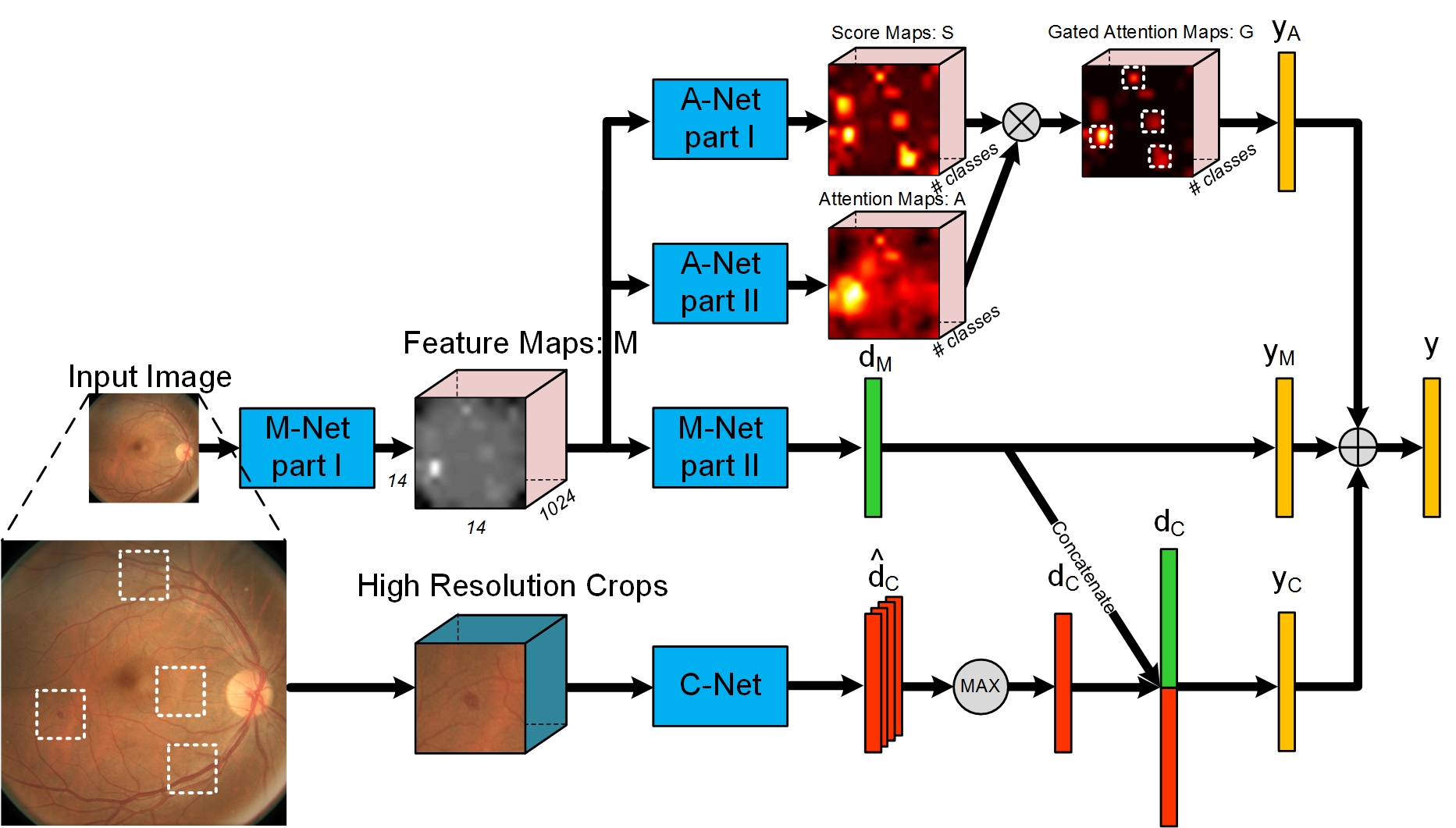}\\
   \caption{An overview of Zoom-in-Net. It consists of three sub-networks. M-Net and C-Net classify the image and high resolution suspicious patches, respectively, while A-Net generates the gated attention maps for localizing suspicious regions and mining lesions. }\label{fig:intro}
\end{figure} 

\section {Architecture of Zoom-in-Net}
The proposed Zoom-in-Net learns from image-level supervisions of DR detection, however equipped with the function to both classify images and localize the suspicious regions.
It mimics the zoom-in process of a clinician examining an image by selecting highly suspicious regions on a high resolution image and makes a decision according to both the global image and local patches.

Our Zoom-in-Net consists of three parts as shown in Fig.\ref{fig:intro}: a main network (M-Net) for DR classification, a sub-network, Attention Network (A-Net), to generate attention maps, and another sub-network,  Crop-Network (C-Net), which takes high resolution patches of highest attention values as input and correct predictions from M-Net. Our illustration is based on the 5-level DR detection task, i.e., 0 - No DR; 1 - Mild; 2 - Moderate; 3 - Severe and 4 - Proliferative DR. It can be easily adapted to different classification tasks.

\subsubsection {Main Network (M-Net)}
The M-Net is a CNN which takes an image as input and processes it by stacks of linear operations including convolutions, batch normalization, non-linear operations like Pooling and Rectified Linear Units. We adopt the Inception-Resnet \cite{szegedy2016inception} as the architecture of M-Net. 
The intermediate feature maps produced by the layer {\it inception\_resnet\_c\_5\_elt}, i.e. $M\in \mathcal{R}^{1024\times 14\times 14}$,  separate the M-Net into two parts as shown in Fig. \ref{fig:intro}. It is followed by a fully connected layer and mapped into a probability vector  $y_M\in \mathcal{R}^5$, which indicates the probability of the image belonging to each disease level. $M$ is further used as input to the Attention Network.

As the Kaggle's challenge provides both left and right eyes of a patient, we also utilize the correlation between two eyes.   Statistics show that more than  $95\%$ of the eye pairs have the scores differ by at most 1. Therefore,  we concatenate the features of both eyes from M-Net together and train the network to take advantage of it in an end-to-end manner.   

\subsubsection {Attention Network (A-Net)} The A-Net takes the feature maps $M$ as input. It consists of two branches. The first branch, A-Net part I,  is a convolution layer with $1\times 1$ kernels. It can be regarded as a linear classifier applied to each pixel and produces score maps $S\in \mathcal{R}^{5\times 14\times 14}$ for the 5 disease levels. 
The second branch, A-Net part II, generates attention gate maps with three convolution layers as shown in Fig. \ref{fig:a-net}. In particular, it produces separate attention gate map for each disease level.  Each attention map is obtained by a spatial softmax operator. 
 Intuitively, the spatial softmax forces the attention values to compete each other and concentrate only on the most informative regions.
Therefore, by regarding the attention map $A\in \mathcal{R}^{5\times 14\times 14}$ as a gate, the output for the A-Net is calculated as 
\begin{equation}
G^l=S^l \otimes A^l \label{eqn2}
\end{equation}
where $G^l$, $S^l$ and $A^l$ are gated feature for A-Net, score map and attention map at $l$-th class, respectively, and $\otimes$ denotes element-wise multiplication. Then we can calculate the final score vector $y_A$ by sum pooling $G$ globally, i.e., $y_A^l=\sum_{i,j} G^l_{i,j} \label{eqn3}$.

\begin{figure}[t]
\centering
  \includegraphics[height=0.85cm]{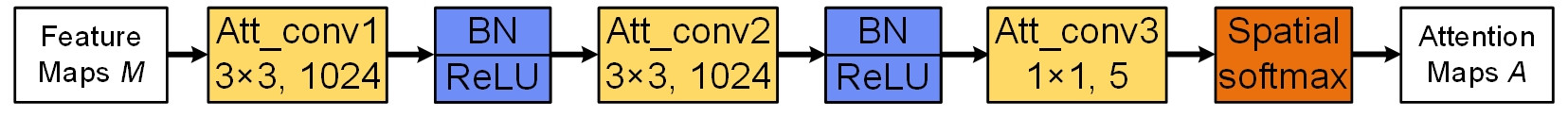}
   \caption{ Structure of A-Net part II. It takes in the feature maps $M$ from M-Net and generates a attention map $A$.  The kernel size and the number of channels are marked at the bottom of the convolution layers. }\label{fig:a-net}
\end{figure} 

\subsubsection{Crop Network (C-Net)}
We further improve the accuracy by zooming-in the suspicious attention regions. 
Specifically, given the gated attention maps $G\in \mathcal{R}^{5\times 14\times 14}$, we first resize it to the same size as the input image. Then we use a greedy algorithm to sample the regions. At each iteration, we record the location of the top response on $G$, and then mask out the $s\times s$ region around it to avoid this region being selected again. We repeat this process until a total of $N$ coordinates are recorded or the maximum attention response is reached.  An example is shown in Fig. \ref{fig:attention}. 

With the recorded locations, we crop the corresponding patches on a higher resolution image for C-Net.  The C-Net has a structure similar to \cite{szegedy2016rethinking}. However, it differs from \cite{szegedy2016rethinking} as it combines features $\hat d_C$ of all patches at layer ``global\_pool''.  Since some patches contain no abnormalities, we apply element-wise max on the feature $\hat d_C$ to extract the most informative feature. This feature is then concatenated to the feature $d_M$ from M-Net and classified by C-Net.

\section{Attention Localization Evaluation and Understanding}
\subsubsection{Attention Localization Evaluation} To verify whether the high response regions contain clues indicating the disease level of the images,  we asked an experienced ophthalmologist to label the lesions of 182 retinal images from EyePACS. The ophthalmologist is asked to draw bounding boxes to tightly cover the lesions related to diabetic retinopathy.  A total of 306 lesions are labeled at last.

We calculate the intersection over Minimum (IoM) between a ground truth box and a sampled box. The sampled boxes are the exactly same 4 boxes used in C-Net.  If IoM is above a threshold, then we consider the sampled box is correct. In this way, we plot two curves of the recall for person and for box V.S. the threshold, respectively, in Table \ref{fig:recall}. The recall for person means that as long as one ground truth box of a person is retrieved by the sampled boxes, we treat this person to be correct. Therefore, it is higher than the recall for box. Note that we achieve a recall of 0.76 and 0.83 at IoM threshold equal to 0.3 for box and person, respectively. This indicates our A-Net can localize the lesions accurately given only image-level supervisions, and partly explains why the C-Net can help improve the predictions made by M-Net.  This is remarkable given that our network is not trained on a single annotated box. We believe increasing the resolution of attention maps ($14\times 14$) could further boosts localization precision.

\subsubsection{Attention Visual Understanding} Furthermore, to better understand the net, we propose a clustering based method to visualize the top responses locations on the gated attention maps.
We partition the features at the same locations on the feature maps $M$ into clusters by the AP clustering algorithm \cite{frey2007clustering}, which is free of a pre-defined cluster number. We can retrieve their corresponding image regions as C-Net input and visualize some of them in Fig. \ref{fig:cluster}.  
Several clusters are discovered with meaningful lesions such as Microaneurysms, blot/frame haemorrhages and hard/soft exudates. This may be very appealing as doctors may find new lesions by examining the clustering results by our method.

\begin{figure}[!tb]\centering
\includegraphics[width=0.95\linewidth]{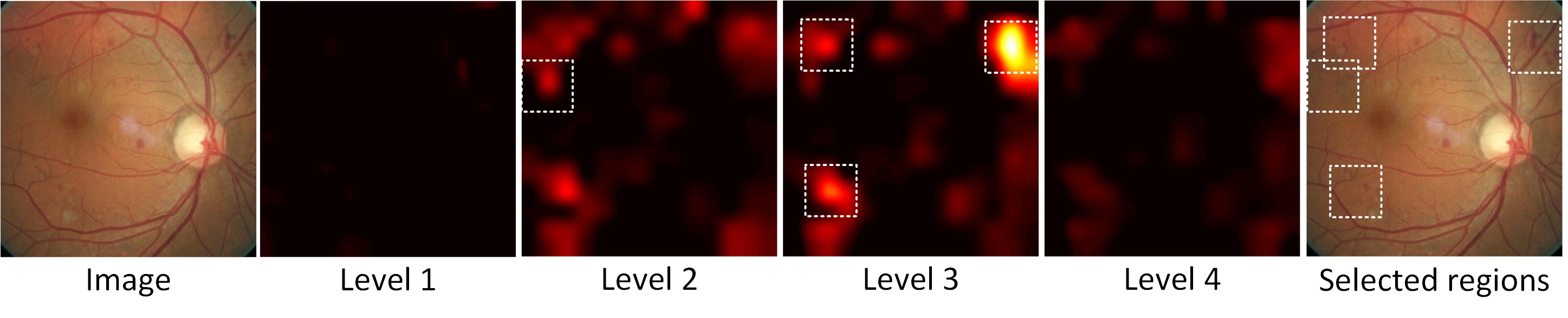}
  \caption{From left to right: image, gated attention maps of level 1-4 and the selected regions of the image. The level 0 gated attention map has no information and is ignored.  }\label{fig:attention}
\end{figure}

\begin{figure}[!tb]\centering
  \includegraphics[width=0.95\linewidth]{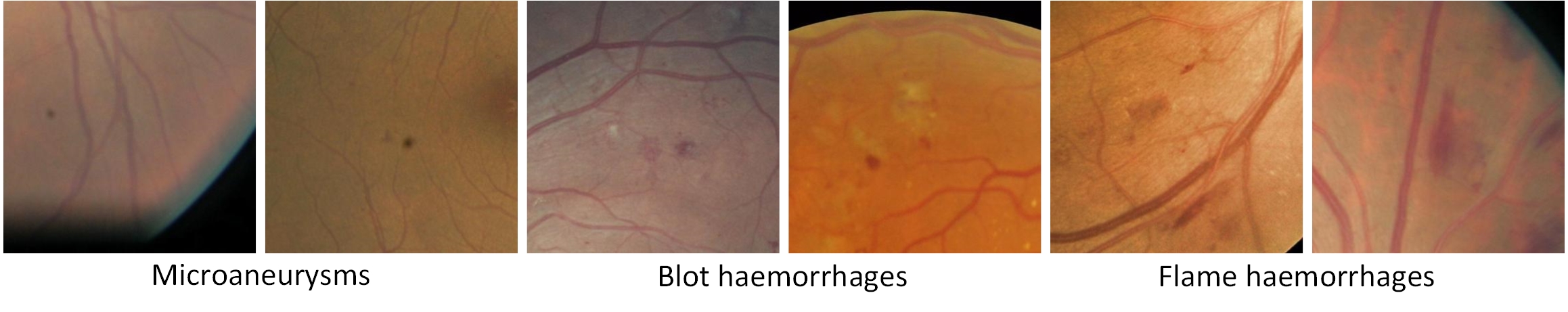}\\
  \includegraphics[width=0.95\linewidth]{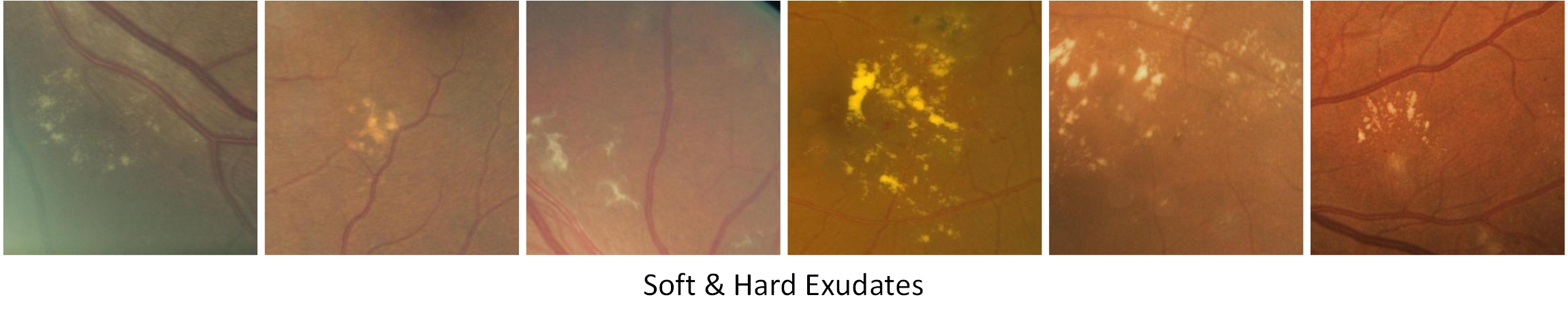}
  \caption{Examples of automatically discovered suspicious regions by clustering features at high respones locations. Some clusters are very meaningful such as microaneurysms, blot haemorrhages, flame haemorrhages, soft exudates and hard exudates.}\label{fig:cluster}
\end{figure}

\section {Quantitative Evaluation}\label{sec:exp}
\subsubsection{Datasets and Evaluation Protocols}
We have evaluated the effectiveness of our Zoom-in-Net on two datasets, the EyePACS and Messidor datasets. 
The Kaggle's Diabetic Retinopathy Detection Challenge (EyePACS) is sponsored by the California Healthcare Foundation. It provides 35k/11k/43k images for train/val/test set respectively, captured under various conditions and various devices. A left and right field is provided for every subject, and a clinician rated the presence of diabetic retinopathy in each image on a scale of 0 to 4. For comparison, we adopt the same official protocol called quadratic weighted kappa score for evaluation. 
The Messidor dataset is a public dataset provided by the Messidor program partners \cite{decenciere2014feedback}.  It consists of 1200 retinal images and for each image, two grades, retinopathy grade, and risk of macular edema, are provided. Only retinopathy grades are used in the present work. 

 \subsubsection{Implemenation details}
The preprocessing includes cropping the images to remove the black borders which contain no information. Data augmentation is done on the training set of EyePACS by random rotations ($0^\circ/90^\circ/180^\circ/270^\circ$) and random flips. 
The training of the proposed Zoom-in-Net includes three phases. We first train M-Net, which is pretrained on ImageNet \cite{russakovsky2015imagenet}, and then train A-Net while fixing the parameters of M-Net.  The C-Net is trained at last together with the other two to obtain the final Zoom-in-Net. During training, we adopt the mini-batch stochastic gradient descent for optimization. We use a gradually decreasing learning rate starting from $10^{-5}$ with a stepsize of 20k and momentum of 0.9. The whole framework is trained with the Caffe library \cite{jia2014caffe}.

\subsubsection{Experiment results on the EyePACS Dataset}

\begin{table}[!tb]
\minipage{0.5\textwidth}
\centering
\includegraphics[width=0.85\linewidth]{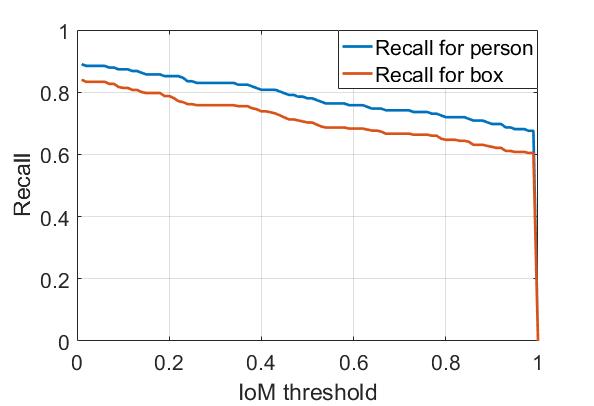}
\centering
\caption{AUC for normal/abnormal}
\label{fig:recall}
\endminipage
\minipage{0.5\textwidth}
\centering
\begin{tabular}{lclcl}
Algorithms        & val set & test set \\\hline
Min-pooling \cite{drdlb}      & 0.86    & 0.849    \\
o$\_$O          & 0.854   & 0.844    \\
Reformed Gamblers & 0.851   & 0.839    \\\hline
M-Net             & 0.832   & 0.825    \\
M-Net+A-Net       & 0.837   & 0.832    \\
Zoom-in-Net   & 0.857   & 0.849    \\
Ensembles         & \textbf{0.865}   & \textbf{0.854}  \\\hline
\end{tabular}
\caption{Comparison to top-3 entries on Kaggle' challenge.}\label{tbl:res}
\endminipage\hfill
\end{table}

We thoroughly evaluate each component of Zoom-in-Net on the EyePACS dataset.  
 As can be seen in Table. \ref{tbl:res}, the M-Net alone achieves $0.832$/$0.825$ on the val/test set, respectively. Adding the branch of A-Net  only improves the score by $0.5\% $ on both sets. This is not surprising as no additional information is added in the A-Net.

Moreover, we use the gated attention maps generated by A-Net to extract suspicious regions and train C-Net.  We observed that on an image resized to $492\times 492$, the area of pathological regions is usually smaller than $200\times 200$. Therefore, we set the region size $s$ to be $200$ and the number of cropped regions $N$ to be 4 throughout the experiments. We cropped $384\times 384$ patches from high resolution images of size $1230\times 1230$ as input of C-Net.  During the training of the complete Zoom-in-Net, one mini-batch contains a pair of whole images and 4 high resolution patches for each image, respectively. It almost reaches the up limit for a Tesla K40 GPU card, so we let  the network update its parameters after every $12$ mini-batches to match the training of M-Net. Finally the proposed Zoom-in-Net achieves the kappa score of $0.857$ and $0.849$ on the two sets, comparable to the first rank entry Min-pooling (0.86/0.849) \cite{drdlb}. With an ensemble of three models, our final results ended up at 0.865/0.854.

\subsubsection{Experiment results on the Messidor Dataset}
To further evaluate the performance, the proposed Zoom-in-Net is applied to the independent dataset Messidor
for DR screening. As Messidor has only 1200 images, the size of which is small to train CNNs, Holly et al.\cite{vo2016new} suggested extracting features from the proposed net trained on other dataset like EyePACS to develop classifiers later. Since Messidor and EyePACS employ different annotation scales (Messidor: 0 to 3, EyePACS: 0 to 4), we follow a protocol similar to \cite{vo2016new} and conduct two binary classification tasks ({referable V.S. non-referable, normal V.S. abnormal}) to realize the evaluation cross dataset and prior studies. 

We extract five dimensional probability feature vectors from the last layer of Zoom-in-Net, and use Support Vector Machines (SVM) with rbf kernal, implemented by the LibSVM library on MATLAB \cite{CC01a}, for binary classification. For referable/nonreferable, Messidor Grade 0 and 1 is considered as nonreferable, while Grade 2 and 3 is defined to be referable to specialists. 10-fold cross-validation on entire Messidor is introduced to be compatible with \cite{vo2016new,Pires2015Beyond}. For normal/abnormal classification, the SVM is trained using extracted features from the training set of EyePACS and tested on entire Messidor. Only images graded 0 on EyePACS/Messidor are assigned as normal, otherwise as abnormal. 

\begin{table}[!tb]
\minipage{0.5\textwidth}
\begin{tabular}{lcc}
\hline
Method  & AUC   & Acc.  \\ \hline
Lesion-based \cite{Pires2015Beyond} & 0.760   & -    \\ 
Fisher Vector \cite{Pires2015Beyond} & 0.863  &  -    \\
VNXK/LGI \cite{vo2016new}      & 0.887  & 0.893  \\
CKML Net/LGI \cite{vo2016new}  & 0.891  & 0.897   \\
Comprehensive CAD \cite{S2011Evaluation} & 0.91 & - \\
Expert A \cite{S2011Evaluation} & 0.94 & - \\
Expert B \cite{S2011Evaluation} & 0.92  & - \\
\textbf{Zoom-in-Net}    & \textbf{0.957} & \textbf{0.911} \\
\hline
\end{tabular}\centering
  \caption{AUC for referable/nonreferable}
\label{table1}
\endminipage\hfill
\minipage{0.5\textwidth}
\begin{tabular}{lcc}
\hline
Method            & AUC   & Acc.  \\ \hline
Splat feature/kNN \cite{Tang2013Splat} & 0.870 &   -    \\ 
VNXK/LGI \cite{vo2016new}          & 0.870 & 0.871 \\
CKML Net/LGI \cite{vo2016new}      & 0.862 & 0.858 \\
Comprehensive CAD \cite{S2011Evaluation} & 0.876 & -       \\
Expert A \cite{S2011Evaluation}          & 0.922 &   -    \\
Expert B \cite{S2011Evaluation}          & 0.865 &   -    \\
\textbf{Zoom-in-Net}        & \textbf{0.921} & \textbf{0.905} \\
\hline
\end{tabular}\centering
\caption{AUC for normal/abnormal}
\label{table2}
\endminipage
\end{table}

The area under the receiver operating curve (AUC) is used to quantify the performance. Table \ref{table1} and \ref{table2} show results of our methods compared with previous studies. To the best of our knowledge, we achieve the highest AUC for both normal and referral classification on Messidor dataset. Zoom-in-Net performs comparably to two experts reported in \cite{S2011Evaluation}. At a specificity of 0.5, the sensitivity of Zoom-in-Net is 0.978 and 0.960, respectively for the normal and referral task. 

\section {Conclusions}
In this work, we proposed a novel framework Zoom-in-Net which achieves state-of-the-art performance on two datasets. Trained with only image-level supervisions, Zoom-in-Net can generate attention maps which highlight the suspicious regions. The localization ability of the gated attention maps is validated and found to be promising.  Further experiments show the high response regions on gated attentions correspond to potential lesions in DR, and thus can be used to further boost of performance for classification.

\bibliographystyle{splncs03}
\bibliography{egbib}
%
\end{document}